# Sufficiency, Separability and Temporal Probabilistic Models

Avi Pfeffer
Division of Engineering and Applied Sciences
Harvard University
avi@eecs.harvard.edu

## Abstract

Suppose we are given the conditional probability of one variable given some other variables. Normally the full joint distribution over the conditioning variables is required to determine the probability of the conditioned variable. Under what circumstances are the marginal distributions over the conditioning variables sufficient to determine the probability of the conditioned variable? Sufficiency in this sense is equivalent to additive separability of the conditional probability distribution. Such separability structure is natural and can be exploited for efficient inference. Separability has a natural generalization to conditional separability.

Separability provides a precise notion of hierarchical decomposition in temporal probabilistic models. Given a system that is decomposed into separable subsystems, exact marginal probabilities over subsystems at future points in time can be computed by propagating marginal subsystem probabilities, rather than complete system joint probabilities. Thus, separability can make exact prediction tractable. However, observations can break separability, so exact monitoring of dynamic systems remains hard.

## 1 Introduction

Bayesian networks (BNs) use conditional independence in order to provide compact representations of probability distributions. This structure also supports efficient inference algorithms. Dynamic Bayesian networks (DBNs) use a similar conditional independence structure in the representation of temporal probabilistic models. However, this structure has proven resistant to exploitation for efficient inference. The problem, of course, is that even when there is local conditional independence in the dynamic model, in the long run all variables become correlated. It would seem therefore, that if we want to perform exact inference, we can do no better than to compute a complete joint probability distribution over the set of state variables at time $t$ (or at least the subset of the state variables that have an effect on the next state).

There is a strong intuition that many dynamic systems can be hierarchically decomposed into weakly interacting subsystems, and such a decomposition should be capable of supporting efficient reasoning about the system. However, previous attempts to represent such decompositions (e.g. [FKP98]) have run into the same inference difficulties as DBNs. Boyen and Koller [BK98, BK99] have shown that hierarchical structure can be exploited for approximate monitoring, but this only applies to approximate inference, and the quality of the approximation is highly sensitive to the numbers in the conditional probability tables and not just the structure. The search for inference-supporting structures in dynamic systems is still on.

This paper identifies such a structure. The key idea is that even if all the state variables become correlated, it may not be necessary to actually compute complete joint distributions over the state. If we are only interested in marginal probabilities of particular variables, can we reason by propagating marginal distributions over subsets of state variables rather than complete joint distributions? What structure would support that? The approach is similar in spirit to Lauritzen's method [Lau92] for inference with conditional Gaussian models. Even though the posterior marginals are not Gaussian, their means and variances can be propagated exactly as if they were.

In studying this question about dynamic systems, I was led to a more specific question about static BNs. Suppose a variable $Y$ has parents $X_1, \ldots, X_n$. Under what circumstances are the $X_i$ *sufficient* for $P(Y \mid X_1, \ldots, X_n)$, in the sense that it is sufficient to know the marginals over the $X_i$ in order to determine $P(Y)$, rather than the full joint over the $X_i$? It turns out that sufficiency is equivalent to additive *separability* of the conditional probability $P(Y \mid X_1, \ldots, X_n)$ into terms that depend on each of the $X_i$ individually. Separability is a natural structure, and is closely related to a form of context-specific independence [BFGK96]. It is quite easy to exploit separability for efficient inference. There is also a natural generalization to a more complex notion of *conditional separability*.

While sufficiency and separability are useful in their own right for static BNs, what about the original question about temporal models? Can we find subsets of state variables that are *self-sufficient*, i.e., that the marginals over these variables at time $t$ are sufficient to obtain correct marginals over them at time $t+1$? It turns out that some temporal systems can indeed be hierarchically decomposed into subsystems in a way that corresponds to separability. As a result, it is indeed possible to predict the marginal distributions over the state of a subsystem by propagating marginal probabilities of the different subsystems.

This is a satisfying result, since it provides a tractable exact solution of one problem in temporal probabilistic inference. However, this is only a limited success, because as soon as observations are introduced, propagating marginals is no longer correct, and therefore the exact monitoring problem is still hard, even for systems that decompose into separable subsystems.

## 2 Sufficiency and Separability

Let us begin with a simple situation. Consider three variables $X$, $Y$, and $Z$, with $Z$ depending on $X$ and $Y$ according to the conditional probability distribution $P(Z \mid XY)$. I will use the notation $\Delta^{XY}$ to denote the space of probability distributions over $XY$, and for $q \in \Delta^{XY}$, $q_X$ denotes the marginal of $q$ over $X$.

If a joint distribution $q$ over $XY$ is given, $P(Z)$ is determined by $\sum_{xy} q(xy) P(Z \mid xy)$. Thus the conditional probability of $Z$ given $XY$ determines a function from $\Delta^{XY}$ to $\Delta^Z$. Normally, this function depends on the joint distribution $q$ over $XY$, but in some cases it is fully determined by the marginals $q_X$ and $q_Y$. This idea leads to the following definition.

**Definition 2.1:** Let $X$, $Y$ and $Z$ be variables, and let $P(Z \mid XY)$ be given. Define the function $\Phi^P : \Delta^{XY} \to \Delta^Z$ by $\Phi^P(q) = \sum_{xy} q(xy) P(Z \mid xy)$. We say that $X$ and $Y$ are *sufficient for $Z$ under $P$* if, for any $q^1, q^2$ in $\Delta^{XY}$ such that $q_X^1 = q_X^2$ and $q_Y^1 = q_Y^2$, $\Phi^P(q^1) = \Phi^P(q^2)$. When $P$ is clear from context, we will simply write $\Phi$ and drop "under $P$". ∎

Sufficiency is a desirable inference property. Suppose we have a BN containing $X$, $Y$ and $Z$ as well as other nodes, and we want to compute $P(Z)$ given some evidence $e$. Suppose that $X$ and $Y$ are the parents of $Z$, and d-separate $Z$ from $e$, but are not themselves conditionally independent given $e$. One way to compute $P(Z)$ is to compute $P(XY \mid e)$ and then apply $\Phi$. However, if $X$ and $Y$ are sufficient for $Z$, we can compute $P(X \mid e)$ and $P(Y \mid e)$ separately, and apply $\Phi$ as if $X$ and $Y$ were independent given $e$. Is there a representational structure corresponding to this inference property? It turns out that sufficiency is equivalent to additive separability of the conditional probability distribution, defined as follows:

**Definition 2.2:** $P(Z \mid XY)$ is *separable* if there exist conditional distributions $P_X(Z \mid X)$ and $P_Y(Z \mid Y)$, and $\gamma \in [0, 1]$, such that $P(Z \mid XY) = \gamma P_X(Z \mid X) + (1 - \gamma) P_Y(Z \mid Y)$. ∎

**Theorem 2.3:** $X$ and $Y$ are sufficient for $Z$ iff $P(Z \mid XY)$ is separable.

**Proof:** "If" is easy. If $P(Z \mid XY)$ is separable,

$$\begin{aligned}\Phi(q) &= \sum_{xy} q(xy) P(Z \mid xy) \\ &= \sum_{xy} q(xy)[\gamma P_X(Z \mid x) + (1-\gamma) P_Y(Z \mid y)] \\ &= \gamma \sum_x (\sum_y q(xy)) P_X(Z \mid x) + \\ &\quad (1-\gamma) \sum_y (\sum_x q(xy)) P_Y(Z \mid y) \\ &= \gamma \sum_x q_X(x) P_X(Z \mid x) + \\ &\quad (1-\gamma) \sum_y q_Y(y) P_Y(Z \mid y).\end{aligned}$$

so $\Phi(q)$ depends only on the marginals $q_X$ and $q_Y$.

For "only if", assume $X$ and $Y$ are sufficient for $Z$, and let $z_1$ be $\arg\max_z (\max_{xy} P(z \mid xy) - \min_{xy} P(z \mid xy))$. I.e., $z_1$ is the value of $Z$ most affected by $XY$. Let $x_1, y_1$ be $\arg\min_{xy} P(z_1 \mid xy)$, and let $P_1 = P(Z \mid x_1 y_1)$. Write $\alpha_i = P(Z \mid x_i y_1) - P_1$, and $\beta_j = P(Z \mid x_1 y_j) - P_1$. Now, for any $i$ and $j$, consider the distributions $q^1$ and $q^2$ where $q^1$ assigns probability $1/4$ to each of the assignments $(x_1, y_1)$, $(x_1, y_j)$, $(x_i, y_1)$, and $(x_i, y_j)$, while $q^2$ assigns probability $1/2$ to each of $(x_1, y_1)$ and $(x_i, y_j)$. Since $q^1$ and $q^2$ have the same marginals, by sufficiency $\Phi(q^1) = \Phi(q^2)$, so $P(Z \mid x_1 y_1) + P(Z \mid x_i y_j) = P(Z \mid x_1 y_j) + P(Z \mid x_i y_1)$, and therefore $P(Z \mid x_i y_j) = P_1 + \alpha_i + \beta_j$.

Write $\alpha^* = \max_i P(z_1 \mid x_i y_1) - P(z_1 \mid x_1 y_1)$ and $\beta^* = \max_j P(z_1 \mid x_1 y_j) - P(z_1 \mid x_1 y_1)$, and set $\gamma = \frac{\alpha^*}{\alpha^* + \beta^*}$. Set $P_X(x_i) = P_1 + \frac{1}{\gamma} \alpha_i$, and $P_Y(y_j) = P_1 + \frac{1}{1-\gamma} \beta_j$. Then $P(Z \mid x_i y_j) = \gamma P_X(Z \mid x_i) + (1-\gamma) P_Y(Z \mid y_j)$, as required. It is not hard to show that $P_X(Z \mid x_i)$ and $P_Y(Z \mid y_j)$ are probability distributions — details are omitted. ∎

Separability has a simple intuitive interpretation. The dependent variable $Z$ is only influenced by one of its parents $X$ or $Y$, but we don't know which. We can imagine a latent variable that determines which of $X$ and $Y$ actually influences $Z$. With probability $\gamma$ the actual parent is $X$, and with probability $1 - \gamma$ it is $Y$.

If $\alpha^* + \beta^* < 1$ (in the notation of the proof), the separation of $P(Z \mid XY)$ is not unique. The remaining probability mass $1 - \alpha^* - \beta^*$ can be divided in any way between $P_X$ and $P_Y$. The choice of $\gamma = \frac{\alpha^*}{\alpha^* + \beta^*}$ is particularly natural. $\alpha^*$ is the maximum possible effect $X$ can have on $Z$ if $Y$ is held fixed, while $\beta^*$ is the maximum effect of $Y$ as $X$ is held fixed, so together $\alpha^*$ and $\beta^*$ can be viewed as determining the relative importance of the two parents.

The above definitions and analysis can be generalized to the case where $Z$ has multiple parents $X_1, \ldots, X_n$. $\Phi$ is now a function from $\Delta^{X_1, \ldots, X_n}$ to $\Delta^Z$ given by $\Phi(q) = \sum_{x_1, \ldots, x_n} q(x_1, \ldots, x_n) P(Z \mid x_1, \ldots, x_n)$. We say that $X_1, \ldots, X_n$ are sufficient for $Z$ if $\Phi(q)$ depends only on the marginals of $q$ over the individual $X_i$. We say that $P(Z \mid X_1, \ldots, X_n)$ is separable if it can be written as $\sum_i \gamma_i P_i(Z \mid X_i)$, where the $P_i$ are conditional distributions and $\sum_i \gamma_i = 1$. Using induction, and similar ideas to the proof of Theorem 2.3, one can show that $X_1, \ldots, X_n$ are sufficient for $Z$ iff $P(Z \mid X_1, \ldots, X_n)$ is separable.

### 2.1 Comparison to Other Structures

Separability is incomparable to conditional independence. $X$ and $Y$ may be conditionally independent given $Z$, but not sufficient for $Z$, and vice versa. However, if $Y$ is conditionally independent from $Z$ given $X$, then trivially $X$ and $Y$ are sufficient for $Z$.

Separability is closely related to *context-specific independence* (CSI) [BFGK96]. In fact, it can be viewed as a case of CSI where the context variable is latent. Separability corresponds to a strong form of CSI, where only one parent is relevant given the context.

Another framework which superficially has some of the same properties as separability is that of *causal independence* [HB94], which includes the well-known noisy-or model. In this framework, one also gets a decomposition of a large conditional probability distribution in terms of individual dependence on each of the parents. However, separability and causal independence are not related; they correspond to two very different ways in which causes can interact. With causal independence, the different causes are all active simultaneously, but their actions happen independently. With separability, exactly one of the different causes is active, but we don't know which. Causally independent models do not have the sufficiency property.

Separability is related to the concept of synergy from the qualitative probabilistic network (QPN) framework [Wel90]. Additive synergy characterizes whether the different causes of an effect reinforce each other or interfere with each other. Separable conditional distributions exhibit zero synergy — the different causes neither reinforce nor interfere with each other, because only one of them is actually active. In contrast, the noisy-or model exhibits negative synergy; if multiple causes are on, some of their work is wasted because another cause would have achieved the effect anyway.

### 2.2 Exploiting Separability in Bayesian Network Inference

Earlier I described how sufficiency can be used to simplify particular BN computations. It is in fact possible to exploit separability in a general way in BN inference. One idea is to make the latent selection variable explicit, and then use methods that have been proposed for exploiting CSI. Two methods are proposed in [BFGK96]. One is to condition on the context variable, with the resulting networks being much simpler. This method is hard to integrate into most BN implementations, which do not use conditioning. The second method is based on a network transformation, which introduces explicit multiplexer nodes to represent models with CSI. In some cases, this can produce a simpler network than the original. However, this transformation is of no benefit in our case, since the selector variable is essentially already acting as a multiplexer; the result node after the transformation will have as many parents as before. Poole and Zhang [Poo97, ZP99] provide a comprehensive method for exploiting CSI in inference, based on an extended version of the variable elimination algorithm that uses either partial functions or rules. While this method could be used for separability, it requires a significant extension to standard BN inference methods.

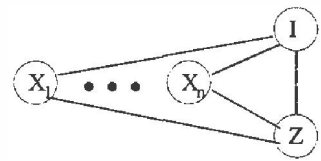

Figure 1: Graph resulting from separability decomposition

As it happens, there is a simple way to exploit separability within the standard variable elimination or junction tree inference frameworks. If $P(Z \mid X_1, \ldots, X_n)$ is separable, it has the form $\sum_i \gamma_i P_i(Z \mid X_i)$. The idea is to turn this summation into a sum-of-products expression, that can then be used in variable elimination. This is achieved by making the selection variable $I$ explicit, and introducing a factor $g_i(I, Z, X_i)$ for $i = 1, \ldots, n$, where

$$g_i(j, z, x_i) = \begin{cases} \gamma_i P_i(z \mid x_i) & \text{if } j = i \\ 1 & \text{otherwise} \end{cases}$$

Now,
$\sum_i \prod_j g_j(i, z, x_j) = \sum_i g_i(i, z, x_i) \prod_{j \neq i} g_j(i, z, x_j) = \sum_i \gamma_i P_i(z \mid x_i) = P_i(z \mid x_1, \ldots, x_n)$. We can therefore replace $P(Z \mid X_1, \ldots, X_n)$ with the sum-of-products expression $\sum_i \prod_j g_j(I, Z, X_j)$. The benefits of this transformation should be clear from looking at the graph corresponding to this expression, shown in Figure 1. The graph is triangulated, and contains $n$ cliques of size 3, as opposed to the clique of size $n$ for $P(Z \mid X_1, \ldots, X_n)$.

This decomposition took advantage of the fact that separability corresponds to a special form of CSI in which there is a single context variable that is used to determine which parent is relevant. In conjunction with the transformation from [BFGK96], it can actually be used as a general method for dealing with CSI within a standard inference algorithm. The transformation shows how to transform any tree-structured conditional probability table into a network with multiplexers. The transformation presented here can then be used to replace the multiplexer CPTs with sum-of-products expressions.

## 3 Conditional Separability

Definitions 2.1 and 2.2 generalize naturally to cases where $\mathbf{X}$ and $\mathbf{Y}$ are sets of variables. If they are disjoint, Theorem 2.3 continues to hold of course, because $\mathbf{X}$ and $\mathbf{Y}$ can be treated as individual variables taking values in the product space. However, if they are not disjoint, Theorem 2.3 fails. It is possible for $\mathbf{X}$ and $\mathbf{Y}$ to be sufficient for $Z$, while $P(Z \mid \mathbf{X} \cup \mathbf{Y})$ is not expressible as $\gamma P_\mathbf{X}(Z \mid \mathbf{X}) + (1-\gamma) P_\mathbf{Y}(Z \mid \mathbf{Y})$. For example, suppose $X, W, Y$ and $Z$ are Boolean variables, and suppose $P(z \mid XWY)$ is 1 if $(x \wedge \bar{w}) \vee (y \wedge w)$ holds, 0 otherwise. Then $\Phi(q) = \sum_{xwy} q(xwy) P(Z \mid xwy) = q_{XW}(x \wedge \bar{w}) + q_{WY}(y \wedge w)$ and therefore $\{X, W\}$ and $\{W, Y\}$ are sufficient for $Z$. However, it can easily be checked that $P(Z \mid XWY)$ cannot be written as $\gamma P_{XW}(Z \mid XW) + (1-\gamma) P_{WY}(Z \mid WY)$. In order to restore a version of Theorem 2.3, we need to introduce the notion of conditional separability. Notation: if $\mathbf{W} \subseteq \mathbf{X}$, $\mathbf{U} = \mathbf{X} - \mathbf{W}$, and we are given $P(Z \mid \mathbf{X})$ and a value $\mathbf{w}$ of $\mathbf{W}$, $P^\mathbf{w}(Z \mid \mathbf{U})$ is the conditional distribution defined by $P^\mathbf{w}(z \mid \mathbf{u}) = P(z \mid \mathbf{uw})$.

**Definition 3.1:** Let $\mathbf{X}$ and $\mathbf{Y}$ be sets of variables, let $Z \notin \mathbf{X} \cup \mathbf{Y}$ be a variable, and let $\mathbf{W} \subseteq \mathbf{X} \cup \mathbf{Y}$. Let $\mathbf{U} = \mathbf{X} - \mathbf{W}$ and $\mathbf{V} = \mathbf{Y} - \mathbf{W}$. $P(Z \mid \mathbf{XY})$ is *conditionally separable* given $\mathbf{W}$ if for every $\mathbf{w}$, there exist conditional distributions $P_\mathbf{U}^\mathbf{w}$ and $P_\mathbf{V}^\mathbf{w}$, and $\gamma^\mathbf{w} \in [0,1]$, such that $P^\mathbf{w}(Z \mid \mathbf{U} \cup \mathbf{V}) = \gamma^\mathbf{w} P_\mathbf{U}^\mathbf{w}(Z \mid \mathbf{U}) + (1-\gamma^\mathbf{w}) P_\mathbf{V}^\mathbf{w}(Z \mid \mathbf{V})$. ∎

In words, $P(Z \mid \mathbf{XY})$ is conditionally separable given $\mathbf{W}$, if for every assignment $\mathbf{w}$, the conditional distribution over $Z$ after conditioning on $\mathbf{w}$ is separable into $\mathbf{X} - \mathbf{W}$ and $\mathbf{Y} - \mathbf{W}$ components. The components and the value of $\gamma$ are allowed to vary with $\mathbf{w}$. Separability is equivalent to conditional separability given $\emptyset$. Also, we can enlarge the conditioning set while preserving conditional separability: if $P(Z \mid \mathbf{XY})$ is conditionally separable given $\mathbf{W}$, and $\mathbf{W} \subseteq \mathbf{W}' \subseteq \mathbf{X} \cup \mathbf{Y}$, then $P(Z \mid \mathbf{XY})$ is also conditionally separable given $\mathbf{W}'$.

**Theorem 3.2:** $\mathbf{X}$ *and* $\mathbf{Y}$ *are sufficient for* $Z$ *iff* $P(Z \mid \mathbf{XY})$ *is conditionally separable given* $\mathbf{W} = \mathbf{X} \cap \mathbf{Y}$.

**Proof:** For "only if", suppose $\mathbf{X}$ and $\mathbf{Y}$ are sufficient for $Z$ under $P$, and consider any $\mathbf{w}$. Suppose that $q^1, q^2 \in \Delta^{\mathbf{UV}}$ have the same marginals over $\mathbf{U}$ and $\mathbf{V}$. Consider $r^1, r^2 \in \Delta^{\mathbf{X} \cup \mathbf{Y}}$ formed from $q^1$ and $q^2$ by forcing $\mathbf{W}$ to equal $\mathbf{w}$ with probability 1. Then $r^1$ and $r^2$ have the same marginals over $\mathbf{X}$ and $\mathbf{Y}$, and also $\Phi^P(r^i) = \Phi^{P^\mathbf{w}}(q^i)$. By sufficiency of $\mathbf{X}$ and $\mathbf{Y}$ under $P$, $\Phi^P(r^1) = \Phi^P(r^2)$, and so $\Phi^{P^\mathbf{w}}(q^1) = \Phi^{P^\mathbf{w}}(q^2)$. Therefore $\mathbf{U}$ and $\mathbf{V}$ are sufficient for $Z$ under $P^\mathbf{w}$, and the result follows from Theorem 2.3.

For "if", by conditional separability,

$$\begin{array}{rcl}
\Phi^P(q) &=& \sum_{\mathbf{uvw}} q(\mathbf{uvw}) P(Z \mid \mathbf{uvw}) \\
&=& \sum_\mathbf{w} q(\mathbf{w}) \sum_\mathbf{uv} q(\mathbf{uv} \mid \mathbf{w}) P^\mathbf{w}(Z \mid \mathbf{uv}) \\
&=& \sum_\mathbf{w} q(\mathbf{w}) \sum_\mathbf{uv} q(\mathbf{uv} \mid \mathbf{w}) \\
& & (\gamma^\mathbf{w} P_\mathbf{U}^\mathbf{w}(Z \mid \mathbf{u}) + (1-\gamma^\mathbf{w}) P_\mathbf{V}^\mathbf{w}(Z \mid \mathbf{v})) \\
&=& \sum_\mathbf{w} q(\mathbf{w}) \quad (\gamma^\mathbf{w} \sum_\mathbf{u} q(\mathbf{u} \mid \mathbf{w}) P_\mathbf{U}^\mathbf{w}(Z \mid \mathbf{u}) + \\
& & (1-\gamma^\mathbf{w}) \sum_\mathbf{v} q(\mathbf{v} \mid \mathbf{w}) P_\mathbf{V}^\mathbf{w}(Z \mid \mathbf{v}))
\end{array}$$

But if $q^1$ and $q^2$ have the same marginals over $\mathbf{X}$ and $\mathbf{Y}$, $q^1(\mathbf{w}) = q^2(\mathbf{w})$, $q^1(\mathbf{u} \mid \mathbf{w}) = q^2(\mathbf{u} \mid \mathbf{w})$ and $q^1(\mathbf{v} \mid \mathbf{w}) = q^2(\mathbf{v} \mid \mathbf{w})$, so $\Phi^P(q^1) = \Phi^P(q^2)$, which gives sufficiency. ∎

Again, we can generalize naturally to multiple sets of parents. If we are given $P(Z \mid \cup_I \mathbf{X}_i)$, then the $\mathbf{X}_i$ are sufficient for $Z$ under $P$ if $\Phi(q)$ depends only on the marginals over the $\mathbf{X}_i$. As for conditional separability, if the intersection $\mathbf{X}_i \cap \mathbf{X}_j$ is the same $\mathbf{W}$ for all $i$ and $j$, then we can simply say that $P(Z \mid \mathbf{X}_1, \ldots, \mathbf{X}_n)$ is conditionally separable given $\mathbf{W}$ if for every $\mathbf{w}$, $P^\mathbf{w}(Z \mid \mathbf{X}_1 - \mathbf{W}, \ldots, \mathbf{X}_n - \mathbf{W})$ is separable.

If the pairs of sets do not have a common intersection, we can define a more complex hierarchical notion of separability as follows. Given a family $\mathbf{X}_1, \ldots, \mathbf{X}_n$ of sets of variables, a *tree representation* of the family is a tree $T$ that contains one leaf for each $\mathbf{X}_i$. For a variable $X \in \cup_i \mathbf{X}_i$, the *location of* $X$ *in* $T$ is the lowest node of $T$ such that all subsets containing $X$ are at or beneath that node. Figure 2 shows an example tree, with each variable shown at its location.

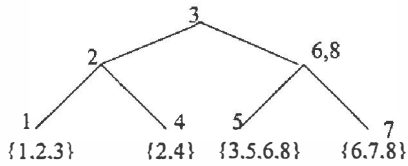

Figure 2: A tree representation

For a given tree representation $T$, let $T_1, \ldots, T_m$ denote the subtrees of $T$. Let $\mathbf{W}$ denote the variables at the root of $T$, and $\mathbf{U}_i$ denote the variables in the subtree $T_i$. We say that $P(Z \mid \cup_i \mathbf{X}_i)$ is $T$-separable if either $T$ is a leaf, or for every assignment $\mathbf{w}$ to $\mathbf{W}$, $P^{\mathbf{w}}(Z \mid \mathbf{U}_1, \ldots, \mathbf{U}_m)$ can be written as $\sum_i \gamma_i^{\mathbf{w}} P_i^{\mathbf{w}}(Z \mid \mathbf{U}_i)$, where $\sum_i \gamma_i^{\mathbf{w}} = 1$, and $P_i^{\mathbf{w}}$ is $T_i$-separable. Using induction, one can show that $\mathbf{X}_1, \ldots, \mathbf{X}_n$ are sufficient for $Z$ iff $P(Z \mid \cup_i \mathbf{X}_i)$ is $T$-separable for any tree representation $T$ of $\mathbf{X}_1, \ldots, \mathbf{X}_n$.

## 4 Separability in Temporal Models

### 4.1 Self-Sufficiency

How is all this relevant for temporal probabilistic models? Let us look at some examples. First, suppose that there are two state variables $X^t$ and $Y^t$. Suppose that in the dynamic model specifying $P(X^t \mid X^{t-1} Y^{t-1})$ and $P(Y^t \mid X^{t-1} Y^{t-1})$, $X^{t-1}$ and $Y^{t-1}$ are sufficient for $X^t$ and also for $Y^t$. The DBN structure is shown in Figure 3. In this dynamic model, the state variables at any point in time are not independent of each other. Nevertheless, because of sufficiency, if we want to know the marginal distributions over $X^t$ and $Y^t$ we only need the marginals over $X^{t-1}$ and $Y^{t-1}$. And similarly, the marginals over $X^t$ and $Y^t$ will give us the marginals over $X^{t+1}$ and $Y^{t+1}$, assuming their are no observations to condition the distribution at time $t$. We can therefore propagate marginals to obtain correct predictions of the marginal probabilities of the state variables at any future time point.

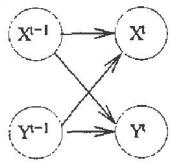

Figure 3: A simple DBN

Now, let us increase the number of state variables to $n$, but still assume that the individual variables $X_1^{t-1}, \ldots, X_n^{t-1}$ are sufficient for each of the $X_1^t, \ldots, X_n^t$. The same situation holds — we only need to propagate marginal distributions over each of the $X_i^t$ to obtain correct predictions of marginals. This is a natural model for a system with a simple information flow, in which at each point in time each variable only receives information from one previous variable, but we don't know which one. We can introduce a matrix $\gamma$, where $\gamma_{ij}$ is the probability that the variable that influences $X_i^t$ is $X_j^{t-1}$. I.e., $\gamma_i$ is the probability distribution over which variable influences $X_i$ at a point in time. Then, associated with each $ij$, there is a model $P_{ij}(X_i^t \mid X_j^{t-1})$ of the particular way in which $X_i$ is influenced by $X_j$.

This type of model is fairly natural for weather systems. The state consists of a variable $X_i$ for each location $i$ in a grid. The variable representing a location may actually be compounded from several variables, such as temperature, pressure and water density. At each point in time, the weather at a location depends stochastically on the weather at one of the neighboring locations at the previous time, but we don't know which neighbor. Climate modelers talk about "packets of air" moving about from one location to another. In this model, $\gamma_i$ encodes the distribution over wind patterns at $i$, while $P_{ij}$ encodes how a packet of air tends to change as it moves from $j$ to $i$.

Let us enrich the model. At any point in time, the wind patterns at different points in a region are not independent, but are correlated by the prevailing wind direction. We can model this by introducing another state variable $W^t$ indicating the prevailing wind direction throughout the entire region at time $t$. This variable will then influence the particular wind pattern at each location, i.e., for each $w$ there is a $\gamma_i^w$. We also make $W^t$ depend on $W^{t-1}$. The DBN is shown in Figure 4. (For convenience, to keep all edges in the model go from one time slice to the next, I have made the wind pattern at location $i$ at time $t$ depend on $W^{t-1}$ instead of $W^t$.) This DBN actually displays little of the structure traditionally sought in DBNs, since each of the $X_i^t$ variables depends on all of the other variables at the previous time. Nevertheless, the seperability makes this a highly structured model, and the structure can be exploited.

It no longer holds that individual state variables at time $t-1$ are sufficient for the state variables at time $t$. However, the doubletons $\{W^{t-1}, X_j^{t-1}\}$ are sufficient for each $X_i^t$. This fact alone is not enough to provide a prediction method via propagating marginals, because we need to maintain marginals over $\{W^t, X_i^t\}$, and not just the individual variables. In fact, however, the sets $\{W^{t-1}, X_j^{t-1}\}$ are sufficient for each $\{W^t, X_i^t\}$. We can see this by checking that for a given value $w$ of $W^{t-1}$, $P^w(W^t, X_i^t \mid \mathbf{X}^{t-1}) = P(W^t \mid w) P^w(X_i^t \mid \mathbf{X}^{t-1}) = \sum_j \gamma_{ij}^w P(W^t \mid w) P_{ij}(X_i^t \mid X_j^{t-1})$,

426 PFEFFER UAI 2001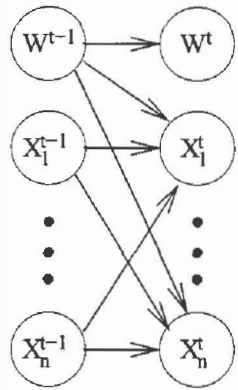

Figure 4: Weather DBN

and therefore $P(W^t, X_i^t \mid \mathbf{X}^{t-1}, W^{t-1})$ is conditionally separable given $W^{t-1}$. As a result, we can propagate marginals over the pairs $(W^t, X_i^t)$, to obtain correct predictions for marginal probabilities over the pairs at any future time.

These examples motivate the definition of a self-sufficient family of sets of variables. For simplicity of presentation, I will assume that all state variables in a DBN depend only on variables from the previous time slice. Of course, real DBN models may be richer, including dependencies within a time slice, but this simplified language captures all we need to develop the theory.

**Definition 4.1:** Consider a DBN with state variables $\mathbf{S}$. A family of subsets $\mathbf{X}_1, \ldots, \mathbf{X}_n$ of $\mathbf{S}$ is *self-sufficient* if $\cup_i \mathbf{X}_i = \mathbf{S}$, and for each $i$, $\mathbf{X}_1^{t-1}, \ldots, \mathbf{X}_n^{t-1}$ are sufficient for $P(\mathbf{X}_i^t \mid \mathbf{S}^{t-1})$. ∎

Given a self-sufficient family, we can define for each $i$ a function $\Phi_i$ from the marginals $q_j^{t-1}$ over the $\mathbf{X}_j^{t-1}$ to a marginal over $\mathbf{X}_i^t$. Then, given an initial distribution $P_0$ over $\mathbf{S}$ and a future time $T$, we can compute the marginals at time $T$ by the following procedure.

For $i = 1$ to $n$
  $q_i^0 = P_0(\mathbf{X}_i)$
For $t = 1$ to $T$
  For $i = 1$ to $n$
    $q_i^t = \Phi_i(q_1^{t-1}, \ldots, q_n^{t-1})$.

It is obvious by induction that this procedure computes the correct marginals at any future time $T$, since $\Phi_i$ always computes the correct marginals at the next time point, given correct marginals at the previous time point. If we have a family with $n$ sets, where the maximum number of variables in a set is $m$, and the maximum number of values of a variable is $b$, the cost of this procedure is $O(Tnb^m)$. In contrast, if the total number of state variables is $M$, the cost of prediction by propagating complete joint distributions is $O(Tb^M)$.

### 4.2 Identifying Self-Sufficient Families

Our goal, then, is to identify self-sufficient families of sets of state variables, in which the individual subsets are small. This was possible in the above examples, but in general it may not be easy to find non-trivial self-sufficient families (the complete set of state variables is of course sufficient for itself).

One might think that a technique based on merging variables into compound variables would work, as it did in the weather example. Such a method might be based on the following rule, which seems plausible: if $\mathbf{X}$ and $\mathbf{Y}$ are sufficient both for $Z_1$ and $Z_2$, and $Z_1$ and $Z_2$ are conditionally independent given $\mathbf{X} \cup \mathbf{Y}$, then $\mathbf{X}$ and $\mathbf{Y}$ are sufficient for $Z = Z_1 \times Z_2$. Unfortunately, this is wrong. Figure 5 shows a simple counterexample. Here, $Z_1$ and $Z_2$ are deterministic copies of $X$ and $Y$ respectively. Obviously $X$ and $Y$ are sufficient for both $Z_1$ and $Z_2$. However they are not sufficient for $Z_1 \times Z_2$. The joint distribution over $Z_1$ and $Z_2$ depends on the joint distribution over $X$ and $Y$, not on their marginals.

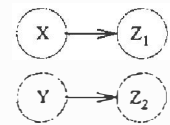

Figure 5: A simple counterexample

A more complex rule for merging variables does hold. If $\mathbf{X}_1$ and $\mathbf{Y}_1$ are sufficient for $Z_1$, and $\mathbf{X}_2$ and $\mathbf{Y}_2$ are sufficient for $Z_2$, then $\mathbf{X}_1 \cup \mathbf{X}_2$, $\mathbf{X}_1 \cup \mathbf{Y}_2$, $\mathbf{Y}_1 \cup \mathbf{X}_2$ and $\mathbf{Y}_1 \cup \mathbf{Y}_2$ are sufficient for $Z = Z_1 \times Z_2$. However, it is not clear how useful this rule is in actually identifying self-sufficient families. Applying it quickly leads to large sets in the family, and normally it will need to be repeatedly applied until there is a set containing all variables, which is useless.

### 4.3 Hierarchical Decomposition

A better approach would be to define ways in which a complex dynamic system can be hierarchically decomposed into separable subsystems, such that the family of subsystems is self-sufficient. We can extend the notion of tree representation defined in Section 3 to capture a hierarchical system decomposition. Recall that a tree representation of a family of sets is a tree containing a leaf for each set. The location of a variable is the lowest node in the tree such that all sets containing the variable are at or beneath the node. A *complete tree representation* is a tree representation that has the added property that for every variable $X$, every leaf beneath the location of $X$ contains $X$. Figure 6(a) shows a complete tree representation. A complete tree representation of a family of subsystems represents a hierarchical decomposition of the system into its subsystems. Variables at the leaves are contained in a single subsystem. Variables at intermediate nodes are shared between a local group of subsystems. Variables at the root are shared among all subsystems.

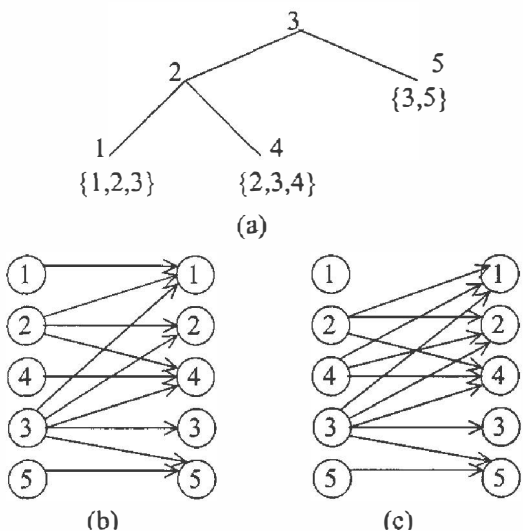

Figure 6: (a) A complete tree representation. (b) Top-down mode. (c) $\{2, 3, 4\}$ takes over one level up.

Suppose that we have a complete tree representation, and we would like the family of subsystems it represents to be self-sufficient. In static BNs, we saw that sufficiency depends on the way information flows from parent to child: $X$ and $Y$ are sufficient for $Z$ if information can only flow from one of $X$ or $Y$ to $Z$ at a time, although we don't know which. A similar effect happens in dynamic systems. There may be many different *modes of operation* in a system, where each mode is characterized by the actual flow of information from variables at the previous time to the current state variables. At any point in time, the system will be in a particular mode, but we may not know what that mode is. In order for self-sufficiency to hold, the different possible modes of operation must all satisfy the property that each subsystem depends on the state of only one subsystem at the previous point in time.

One mode that sastifies this property is a *top-down mode*. In top-down operation, a variable can only be influenced by variables at the same node or its ancestors in the complete tree representation. Figure 6(b) shows a top-down information flow corresponding to the representation of Figure 6(a). Since variables at a node and its ancestors are all contained in the same subsystem, this ensures that each subsystem will only depend on its own state at the previous time. Correlations between subsystems are induced via shared variables. Variables that are shared between subsystems cannot be influenced by variables in individual subsystems.

However, there are other possible modes of operation in which shared variables are influenced by lower level variables. The basic rule is that if a higher level variable $X$ is influenced by variables within a subsystem $S$, then $S$ takes over and influences all subsystems sharing $X$. In particular, another subsystem $S'$ sharing $X$ will not depend on its own previous state, but on that of $S$. Such a mode of operation is called a *take-over mode*. Taking over can happen at any level of the hierarchy. At one extreme, a catastrophic event in one subsystem can influence all other subsystems. More commonly, a subsystem will take over the level above it in the hierarchy, so that it influences the neighboring subsystems. Figure 6(c) shows an example of a take-over mode in which the set $\{2, 3, 4\}$ has taken over the level above it in the hierarchy. As a result, variables 1 and 2 now depend on variable 4, but variable 1 no longer depends on its own previous state. The root of the hierarchy, variable 3, is not taken over, and the operation in the remainder of the hierarchy is top down.

The mode of operation at a particular point in the hierarchy could be determined by variables higher up. For example, the value of variable 3 could determine whether the mode is that of Figure 6(b) or Figure 6(c). Self-sufficiency allows for some quite rich models. On the other hand, some modes of operation are ruled out by self-sufficiency. In particular, a subsystem cannot depend both on its own internal state and on that of another subsystem at the same time. This rules out some traditional notions of weak interaction between subsystems, in which the state of a subsystem is almost completely determined by its own previous state, but may be perturbed by another subsystem. Separability corresponds to a very different kind of decomposition, in which there is a switch determining what influences a subsystem.

### 4.4 Observations

I have shown how self-sufficient families allow us to obtain exact predictions of marginals by propagating marginals. What about monitoring, where we want to maintain the distribution over the current state at each point in time, taking into account observations obtained at each time point? Unfortunately, observations tend to break sufficiency. The problem is that even if we have the correct marginals over a family of subsets of variables, we do not have the joint distribution over all the variables. If an observed variable appears in one subset but not others, it should still condition variables in the other subsets if the variables are not independent in the joint distribution. The marginal distributions do not allow us to perform this conditioning. Therefore, the posterior marginals after conditioning on the observed variable will be wrong.

All is not lost. If an observed variable appears in all subsets in a family, then we can correctly condition in each of the subsets to obtain correct posterior marginals. If a system has a small number of observed variables, and we can contrive to place these variables in the roots of the hierarchical decomposition while maintaining self-sufficiency, we will be able to perform exact monitoring of marginals by propagating marginals. Unfortunately I do not expect this situation to obtain all that often.

## 5 Conclusions and Speculations

In this paper, I have analyzed a desirable inference property — sufficiency — and shown that it is equivalent to a representational structure — separability. The analysis extends to more complex notions of sufficiency and the corresponding structure of conditional separability. I have shown how to exploit separability and conditional separability within the context of Bayesian network inference algorithms.

For temporal probabilistic models, I have shown that some dynamic systems can be decomposed into a self-sufficient family of subsystems, that allows for exact prediction of marginal probabilities without propagating complete joint distributions. This is satisfying, since as far as I know it is the first non-trivial result of its kind. At the same time, the fact that it does not carry over to monitoring is somewhat disappointing.

The results of this investigation were surprising to me. I began with an intuition that some type of hierarchical decomposition would lead to the ability to propagate marginals exactly. I expected the structure to correspond to some sort of traditional notion of weak interaction, based on local independence or near-independence of sets of variables. It turned out that a very different kind of separability structure was needed, corresponding to simple information flow between different subsystems. Systems that have little or no independence, where a variable can depend on many other variables, may still exhibit separability, like in the weather model.

As a result, I believe that studying the information flow in dynamic systems may prove fruitful even when the system is not completely separable. Can separability analysis be integrated with the Boyen-Koller analysis for monitoring? Can the type of information-flow decomposition leading to separability be combined with other notions of weak interaction to provide for near-separability? If so, how can that be exploited for approximate inference?

Separability may be particularly useful in object-based models in which relationships between objects may vary over time. For example, consider a model of a building and the people in it. At any moment, a person is only in one room, but we may not know which. Are there languages that facilitate the definition and identification of separable models?

Finally, it would be interesting to see if notions of separability can be useful in decision making frameworks like factored Markov Decision Processes. Like DBNs, these have proven resistant to being exploited for efficient solution. It seems that more structure than a factored representation is needed to make solving an MDP tractable. Perhaps separability could provide a clue to finding such a structure.

## References

[BFGK96] C. Boutilier, N. Friedman, M. Goldszmidt, and D. Koller. Context-specific independence in Bayesian networks. In *UAI*, 1996.

[BK98] X. Boyen and D. Koller. Tractable inference for complex stochastic processes. In *UAI*, 1998.

[BK99] X. Boyen and D. Koller. Exploiting the architecture of dynamic systems. In *AAAI*, 1999.

[FKP98] N. Friedman, D. Koller, and A. Pfeffer. Structured representation of complex stochastic systems. In *AAAI*, 1998.

[HB94] D. Heckerman and J.S. Breese. A new look at causal independence. Technical report, Microsoft Research MSR-TR-94-08, 1994.

[Lau92] S.L. Lauritzen. Propagation of probabilities, means and variances in mixed graphical association models. *Journal of the American Statistical Association*, 87(420):1098–1108, 1992.

[Poo97] D. Poole. Probabilistic partial evaluation: Exploiting rule structure in probabilistic inference. In *IJCAI*, 1997.

[Wel90] M. Wellman. Fundamental concepts of qualitative probabilistic networks. *Artificial Intelligence*, 44(3):257–303, 1990.

[ZP99] N.L. Zhang and D. Poole. On the role of context-specific independence in probabilistic reasoning. In *IJCAI*, 1999.